\documentclass[letterpaper]{article} % DO NOT CHANGE THIS
\usepackage[]{aaai25}  % DO NOT CHANGE THIS
\nocopyright
\usepackage{times}  % DO NOT CHANGE THIS
\usepackage{helvet}  % DO NOT CHANGE THIS
\usepackage{courier}  % DO NOT CHANGE THIS
\usepackage[hyphens]{url}  % DO NOT CHANGE THIS
\usepackage{graphicx} % DO NOT CHANGE THIS
\usepackage{natbib}  % DO NOT CHANGE THIS AND DO NOT ADD ANY OPTIONS TO IT
\usepackage{caption} % DO NOT CHANGE THIS AND DO NOT ADD ANY OPTIONS TO IT
\usepackage{amsmath}
\usepackage{amssymb}
\usepackage{algorithm}
\usepackage{algorithmic}
\usepackage{colortbl}
\usepackage{xcolor}
\usepackage{subcaption}
\usepackage{multirow}
\usepackage{multicol}
\usepackage{booktabs}
\definecolor{commentcolor}{RGB}{110,154,155}   % define comment color
\newcommand{\PyComment}[1]{\ttfamily\textcolor{commentcolor}{\# #1}}  % add a "#" before the input text "#1"
\newcommand{\PyCode}[1]{\ttfamily\textcolor{black}{#1}} % \ttfamily is the code font
\usepackage{newfloat}
\usepackage{listings}
\DeclareCaptionStyle{ruled}{labelfont=normalfont,labelsep=colon,strut=off} % DO NOT CHANGE THIS
\floatstyle{ruled}
\newfloat{listing}{tb}{lst}{}
\floatname{listing}{Listing}
\title{Vote\&Mix: Plug-and-Play Token Reduction for Efficient Vision Transformer} 
\author{
    Shuai Peng\textsuperscript{\rm 1}, Di Fu\textsuperscript{\rm 2}, Baole Wei\textsuperscript{\rm 1}, Yong Cao\textsuperscript{\rm 2}, Liangcai Gao\textsuperscript{\rm 1}, Zhi Tang\textsuperscript{\rm 1}
}
\affiliations{
    \textsuperscript{\rm 1}Peking University\\
    \textsuperscript{\rm 2}ByteDance\\
    pengshuaipku@pku.edu.cn, 
    fudi.01@bytedance.com,
    baolewei@pku.edu.cn,
    yongc@bytedance.com,
    gaoliangcai@pku.edu.cn,
    tangzhi@pku.edu.cn
}

\usepackage{bibentry}
\begin{document}

\maketitle

\begin{abstract}
 Despite the remarkable success of Vision Transformers (ViTs) in various visual tasks, they are often hindered by substantial computational cost. In this work, we introduce Vote\&Mix (\textbf{VoMix}), a plug-and-play and parameter-free token reduction method, which can be readily applied to off-the-shelf ViT models \textit{without any training}. VoMix tackles the computational redundancy of ViTs by identifying tokens with high homogeneity through a layer-wise token similarity voting mechanism. Subsequently, the selected tokens are mixed into the retained set, thereby preserving visual information. Experiments demonstrate VoMix significantly improves the speed-accuracy tradeoff of ViTs on both images and videos. Without any training, VoMix achieves a 2$\times$ increase in throughput of existing ViT-H on ImageNet-1K and a 2.4$\times$ increase in throughput of existing ViT-L on Kinetics-400 video dataset, with a mere 0.3\% drop in top-1 accuracy. 
\end{abstract}


\section{Introduction}
\label{sec:intro}

Since the migration from Natural Language Processing (NLP) to Computer Vision (CV), Transformers have set new performance benchmarks in a variety of tasks including image classification \cite{dosovitskiy2020image,jiang2021all,liu2021swin,wang2021pyramid} and action recognition \cite{bertasius2021space,feichtenhofer2022masked}, surpassing Convolutional Neural Networks. However, a notable challenge of Vision Transformers (ViTs) lies in their substantial computational cost. This is primarily due to the self-attention mechanism, where the computational cost grows quadratically with respect to the number of tokens. Moreover, maintaining a constant token count across all layers of ViT exacerbates this issue, limiting its applicability in many real-world scenarios.

Recent studies \cite{he2022masked,feichtenhofer2022masked,tong2022videomae,wang2023videomae} have revealed that, compared to languages, visual data exhibits significantly heavy redundancy. A large proportion of tokens within ViT can be discarded and recovered by neighboring tokens. Motivated by it, an acceleration strategy for ViT has emerged, referred to as \textit{token reduction} \cite{haurum2023tokens}, which mitigates computational cost by reducing token number in ViT.

\begin{figure}
    \centering
    \includegraphics[scale=0.3]{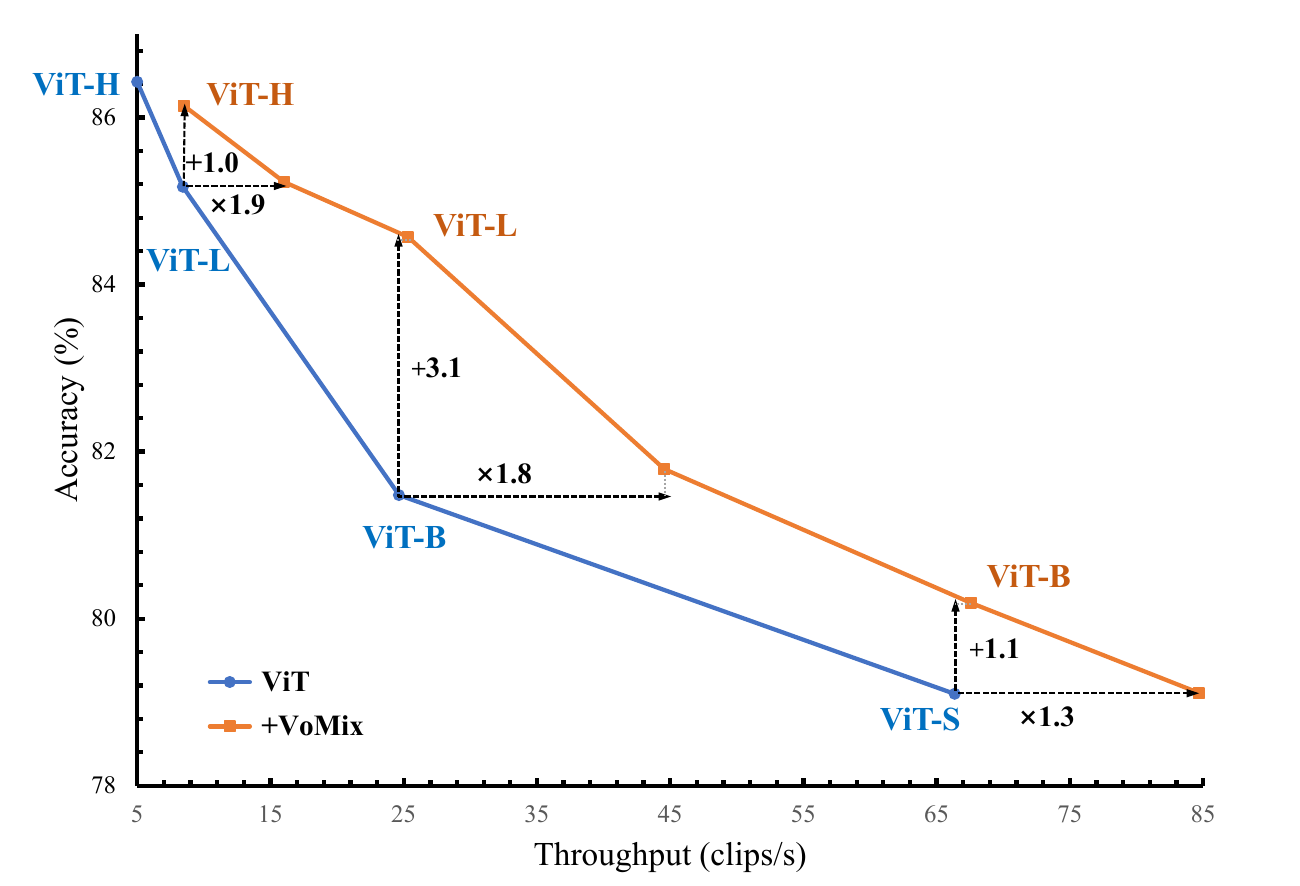}
    \caption{VoMix improves the speed-accuracy tradeoff of ViTs on Kinetics-400.}
    \label{fig:intro_performance}
    \vspace{-0.2in}
\end{figure}

However, there are notable limitations in existing token reduction methods. Some rely heavily on specific tokens (typically class tokens) to assign significance scores to other tokens \cite{fayyaz2022adaptive,yin2022vit}, thus confining their application to particular models only. Some methods introduce extra parameters \cite{kong2022spvit,rao2021dynamicvit}, with the need for model retraining. These drawbacks limit their practical applicability, making adapting token reduction methods to a trained ViT model troublesome.

Recent research \cite{park2022self, long2023bat} has suggested that the attention mechanism in ViTs tends to collapse into homogeneity, where different query tokens elicit identical attention signals. Inspired by this, we argue that tokens with high homogeneity can be more effectively represented by other tokens. Hence, diverging from previous token reduction strategies that focus on discarding insignificant tokens, we aim to reduce token homogeneity. Accordingly, pruning homogenized tokens enhance the efficiency of token utilization in ViT, thereby boosting performance.

Therefore, we introduce Vote\&Mix (\textbf{VoMix}), a plug-and-play, parameter-free token reduction method. In each layer, VoMix identifies tokens with high homogeneity through a \textbf{\textit{voting}} mechanism and then \textbf{\textit{mixes}} them into the retained tokens. Remarkably, VoMix can be applied to off-the-shelf ViT models \textit{without any training}, significantly accelerating inference while maintaining accuracy. Experiments on both image and video datasets, including ImageNet-1K \cite{deng2009imagenet}, Kinetics-400 \cite{kay2017kinetics}, and SSV2 \cite{goyal2017something} demonstrate that VoMix achieves a state-of-the-art tradeoff between computational cost and accuracy. As is shown in Figure \ref{fig:intro_performance}, VoMix significantly improves the speed-accuracy tradeoff of ViT. VoMix achieves improved accuracy at the same speed, and greater speed at the same accuracy. Furthermore, we visually explore VoMix's tendency to retain and mix tokens, discovering that VoMix functions similarly to soft token clustering, thereby accelerating inference while maintaining accuracy. We conduct ablation studies and demonstrate the superiority of the voting mechanism. Finally, we discuss the pruning schedules and acceleration effects of training VoMix.

Compared to other token reduction methods, in addition to the excellent performance, VoMix possesses advantages in the following aspects:

\begin{itemize}
    \item \textbf{Plug-and-Play}: VoMix saves the time and cost for retraining and deployment.
    \item \textbf{Simplicity and Efficiency}: VoMix is a parameter-free method introducing very low computational complexity and allows for flexible model scaling.
    \item \textbf{Broad Applicability}: It can be applied to most mainstream ViTs and excels in image and video modalities.
 \end{itemize}

\section{Related Work}
\label{sec:related}

\subsection{Efficient Vision Transformers}

Since the advent of the Transformer \cite{vaswani2017attention} and its subsequent adaptation in the Vision Transformer \cite{dosovitskiy2020image}, there has been a surge in research aimed at enhancing the efficiency of Transformer models, particularly in the computer vision domain. These include model pruning \cite{chavan2022vision,chen2021chasing,meng2022adavit,song2022cp}, quantization \cite{li2022q,lin2021fq} and efficient attention \cite{shen2021efficient,dao2022flashattention,bolya2022hydra}. Since Transformer allows variable token length, there emerges \textit{token reduction}. It aims to enhance the efficiency of ViT by reducing the number of tokens processed. The proposed method in our paper falls into this category. 

\subsection{Token Reduction}

The prior work on token reduction can be divided into token pruning, token clustering and token merging.

\textbf{Token pruning} reduces tokens by removing less important ones. 
% Pruning-based methods select tokens by assigning significance scores, thereby enhancing model efficiency with minimal information loss.
One typical strategy \cite{fayyaz2022adaptive,yin2022vit} leverages the attention weights of class tokens to estimate per-token keep probabilities. However, the absence of meaningful class tokens in many ViTs limits the applicability. Another strategy \cite{kong2022spvit,rao2021dynamicvit, wei2023tps} employs a learnable module to predict per-token significance scores. While it introduces extra parameters and computational cost, it also requires retraining the model. Inherently, token pruning risks information loss, and score-based sampling strategies tend to discard tokens within the same category, leaving redundancy in others \cite{marin2023token}. Contrary to pruning-based methods, our proposed method focuses on reducing token homogeneity while preserving the information of pruned tokens.

\textbf{Token clustering} reduces tokens by clustering tokens into several clusters. It can be divided into hard-clustering and soft-clustering according to the strategy. Hard-clustering methods \cite{marin2023token,xu2022groupvit,zeng2022not} typically use commonly known clustering methods like K-Means or K-Medoids, and combine tokens within clusters \cite{xu2022groupvit}. These methods often require multiple iterations for clustering and lack flexibility. Soft-clustering methods \cite{zong2022self,renggli2022learning} generally involve parameterized learners to predict cluster centers and assignment matrix, thereby introducing extra parameters. Our proposed method enables efficient token mixture in a soft manner, and no need for training.

\textbf{Token merging} reduces tokens by merging redundant tokens into one. A typical method is ToMe \cite{bolya2022tome}, which gradually merges similar token pairs. The following work \cite{tofu} updates naive average merging to normalized average. Nevertheless, these methods still rely on simply calculating pairwise similarity to select tokens to merge, while neglecting the global homogeneity of the tokens. In contrast, our proposed method offers two key improvements: (1) Voting mechanism: VoMix uses a global voting method to select the most homogeneous tokens. We will demonstrate the effectiveness of voting in the ablation study. (2) Token mix: VoMix performs query fusion within the attention mechanism before applying qkv-attention, which is softer and reduces the time complexity of self-attention to $O(N^2D(1-r))$.

\begin{figure*}[t]
    \centering
    \includegraphics[scale=0.3]{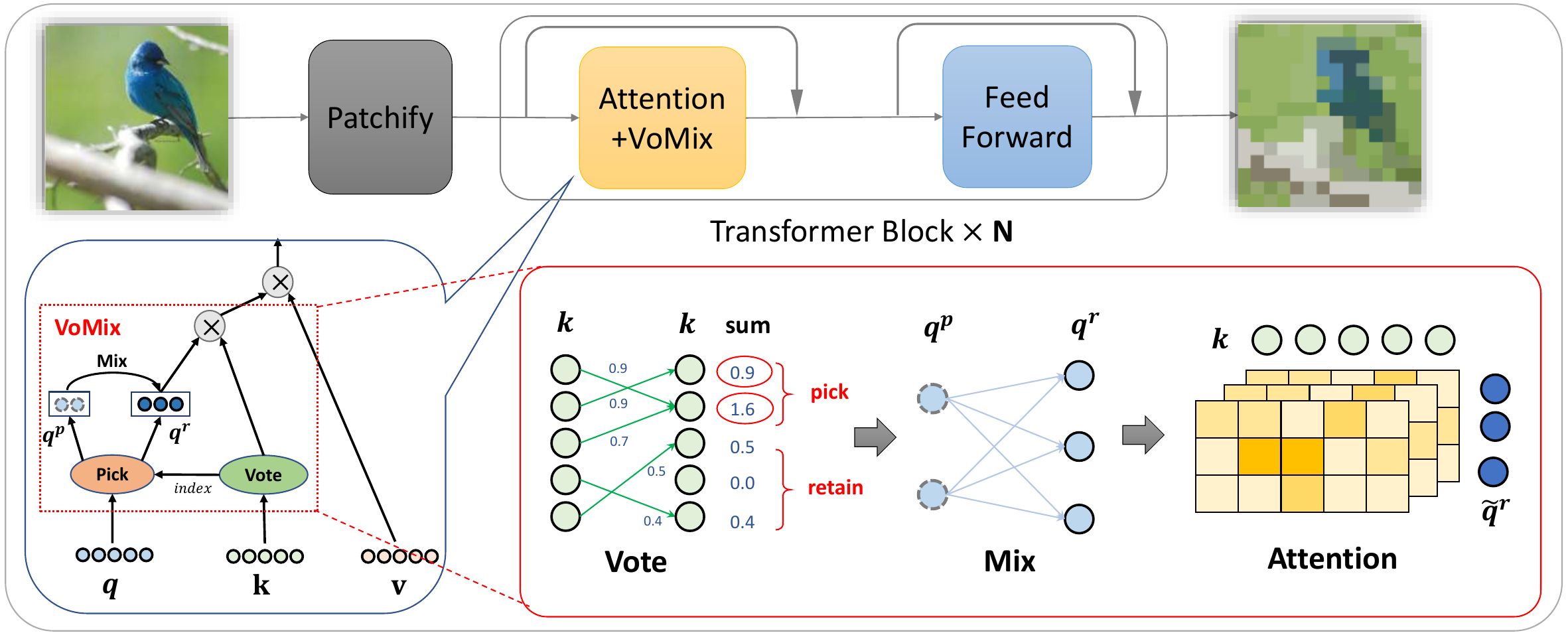}
    \caption{The overview of VoMix. VoMix is a plug-and-play module that can be easily applied to off-the-shelf ViT models. In each transformer block, VoMix reduces a proportion of $r$ tokens in the modified attention mechanism. VoMix has three stages: (1) Vote. VoMix votes $N \cdot r$ tokens out of $N$ tokens via similarity between keys. (2) Mix. VoMix mixes queries of selected tokens into the retained. (3) Attention. VoMix conducts attention using mixed queries and vanilla keys.}
    \label{fig:architecture}
    \vspace{-0.2in}
\end{figure*}

\section{Vote\&Mix}
\label{sec:method}

We introduce VoMix, which alters only the self-attention mechanism in ViT. At each layer, with an initial token count of $N$, VoMix first selects $N \cdot r$ tokens with high homogeneity via token voting. Subsequently, VoMix mixes the selected queries (\textbf{q}) into the retained ones. In the attention mechanism, the mixed $N \cdot r$ queries interact with the original $N$ keys (\textbf{k}) and values (\textbf{v}), ultimately yielding an output of $N \cdot (1-r)$ tokens. Figure \ref{fig:architecture} illustrates the VoMix process.

\begin{algorithm}[t]
\scriptsize
    \PyComment{Input:} \\
    \PyComment{x: token embedding of size (b, n, d)} \\
    \PyComment{r: pruning ratio} \\
    \PyComment{s: mixed size from last layer} \\
    \ \\
    \PyComment{Token Vote} \\
    \PyCode{q,k,v = qkv(x)} \PyComment{(b, h, n, d/h)} \\
    \PyCode{k\_ = k.mean(1)} \PyComment{(b, n, d/h)} \\
    \PyComment{Set the diagonal to -inf} \\
    \PyCode{A = mask\_diag(sim(k\_, k\_))} \PyComment{(b, n, n)} \\
    \PyCode{v\_w, v\_i = A.max(1)} \\
    \PyComment{vote counting} \\
    \PyCode{score = score.scatter\_add(-1, v\_i, v\_w)} \PyComment{(b, n)} \\
    \PyComment{retained index: N*(1-r)} \\
    \PyCode{r\_id = score.argsort(-1)[:,:N*(1-r)]} \\
    \PyComment{pruned index: N*r} \\
    \PyCode{p\_id = score.argsort(-1)[:,N*(1-r):]} \\
    \ \\
    \PyComment{Token Mix} \\
    \PyComment{compute mixture weight from p to r} \\
    \PyCode{W = softmax(A[:, p\_id, r\_id])} \PyComment{(b, nr, n(1-r))}\\
    \PyComment{Query Mix} \\
    \PyCode{q\_w = q.view(b, n, d) * s} \\
    \PyCode{q\_w[:,r\_id,:]+=bmm(W.T, q\_w[:,p\_id,:])} \PyComment{(b, n(1-r), d)} \\
    \PyComment{mix size} \\
    \PyCode{s\_new = s[:,r\_id]+bmm(W.T, s[:,p\_id])} \PyComment{(b, n(1-r))} \\
    \PyComment{scale to orignial size} \\
    \PyCode{q\_new = (q\_w / s\_new).view(b,h,n*(1-r),d/h)} \\
    \PyComment{Attention Mix} \\
    \PyCode{attn = (q\_new*scale)@k.T + log(s)} \\
    \PyCode{x\_new = proj((attn@v).view(b,n*(1-r),d)} \\
    \ \\
    \PyCode{return x\_new, s\_new, x[:,r\_id,:]}
    
\caption{\small PyTorch-style Pseudocode of VoMix.}
\label{algo:vomix-pytorch}
\end{algorithm}

\subsection{Token Vote}

\textbf{Objective:} in the \(l\)-th layer, given the input tokens $X^{l} = \{x^l_1, x^l_2, ..., x^l_N\}$, token voting aims to select a subset $P^{l}$ consisting of $N \cdot r$ tokens with the highest homogeneity, where $r \in [0, 1)$ is the pruning ratio.

Intuitively, a token with high homogeneity implies a high similarity with many other tokens. We adopt a similarity voting strategy to identify these tokens. 

\textbf{Similarity Measurement} Within each transformer block of the ViT, VoMix measures the cosine similarity between tokens, yielding a similarity score matrix $\textbf{A} \in \mathbb{R}^{N \times N}$. Here, we use the head-wise mean of keys (\textbf{k}) as the metric to reduce the additional computation. Mathematically,

\begin{align}
\small
    &\bar{\textbf{k}}_i = \frac{1}{H}\sum_{h=0}^{H}{\textbf{k}_{h,i}}, \quad i \in [1, N]\\
    &\textbf{A}_{i,j} = \frac{\bar{\textbf{k}}_i \cdot \bar{\textbf{k}}_j}{||\bar{\textbf{k}}_i||\cdot||\bar{\textbf{k}}_j||}, \quad i, j \in [1, N]
\end{align}

where $H$ is attention head number and $\textbf{A}_{i,j}$ denotes the cosine similarity between token $i$ and $j$. $\textbf{A}_{i,i}$ is set to $-\infty$ to prevent self-voting.
  
\textbf{Vote Counting.} Each token casts its vote for the most similar token to itself, where the votes are weighted by similarity scores. The score of each token $score$ is the sum of weighted votes received:

\begin{align}
\small
    &z(i) = \arg\max_{j} \textbf{A}_i, \quad i \in [1, N] \\
    &\textit{score}_i = \sum_{j=0}^{N} {\textbf{A}_{j,z(j)} \cdot \delta_{i, z(j)}}, \quad i \in [1, N]
\end{align}

where $z(i)$ denotes the index that token $i$ vote to, $\delta_{a,b}$ is the Kronecker delta, which is 1 if a=b and 0 otherwise. After that, VoMix sorts tokens by $score$, selecting the top $r$ proportion of tokens, as $P^{l}$. The remains form the set $R^{l}$.

\subsection{Token Mix}

\textbf{Objective:} given the selected subset $P^l$ and remained $R^l$, token mixing aims to integrate the tokens of $P^l$ into $R^l$ to preserve the information of $P^l$.

Directly discarding the selected tokens would invariably loss information. To mitigate it, VoMix mixes them into the retained pool. The steps are as follows:

\textbf{Mixture Weight} VoMix gathers the similarity score $\textbf{A}^{'} \in \mathbb{R}^{Nr \times N(1-r)}$ directly from $\textbf{A}$, as the similarity between $P^l$ and $R^l$ . Then the mixture weight $\textbf{W}$ is the softmaxed gathered score $\textbf{A}^{'}$.

\textbf{Query Mix} Query mix conducts a soft feature mixture for queries. Before attention, queries $\textbf{q}^p$ from $P^l$ are mixed into queries $\textbf{q}^r$ from $R^l$ with the mixture weights $\textbf{W}$. Note that token mixing assigns tokens with variable weights, the query $\textbf{q}_i$ needs to be scaled by a mixed size $\textbf{s}_i^{l-1}$ first:

\begin{equation}
\small
    \tilde{\textbf{q}}_i^r = \textbf{q}_i^r \textbf{s}_i^{l-1} + \sum_{j=0}^{N \cdot r} {\textbf{W}_{j,i} \textbf{q}_j^p \textbf{s}_j^{l-1}}, \quad i \in [1, N(1-r)]
\end{equation}

where $\textbf{s}_i^l$ is the mixed size of token $i$ in the $l$-th layer, indicating how many tokens have been mixed into token $i$. The initial size of $\textbf{s}_i^1$ is $1$. Then we update the new weighted size $\textbf{s}_i^l$ and normalize the final query $\tilde{\textbf{q}}_i^r$:

\begin{align}
\small
    &\textbf{s}_i^{l} = \textbf{s}_i^{l-1} + \sum_{j=0}^{N \cdot r} {\textbf{W}_{j,i} \textbf{s}_j^{l-1}}, \quad i \in [1, N(1-r)] \\
    &\tilde{\textbf{q}}_i^r = \tilde{\textbf{q}}_i^r / \textbf{s}_i^{l}, \quad i \in [1, N(1-r)]
\end{align}

 After that, we obtain the mixed queries $\tilde{\textbf{q}}^r \in \mathbb{R}^{N \cdot (1-r)}$.

\textbf{Attention Mix} We conduct self-attention using the mixed queries $\tilde{\textbf{q}}^r$ with original keys $\textbf{k}$ and values $\textbf{v}$. We use proportional attention to pay more attention to larger weighted keys, formulated as:

\begin{equation}
\small
    \text{Attention} = \text{softmax}(\frac{\tilde{\textbf{q}}^r \textbf{k}^{T}}{\sqrt{d}} + \log \textbf{s}^{l-1}) \textbf{v}
\end{equation}

Since $\textbf{k}$ are not mixed in $l$-th layer, we use the size $\textbf{s}^{l-1}$. Finally, we obtain the output tokens $X^l_{out}$ of layer $l$. The pseudocode in Algorithm \ref{algo:vomix-pytorch} shows how VoMix works in pytorch-style pseudocode.

\begin{table*}[t]
    \centering
    \vspace{-0.1in}
    \begin{tabular}{lllll}
        \toprule
        Model & Resolution & Acc & GFLOPs & im/s \\
        \midrule
        ViT-B$^{\text{MAE}}$ & 224 & 83.6 & 17.6 & 304\\
        \textbf{VoMix-B}$^{\text{MAE}}_{r=(5\%)^{12}}$ & 224 & 83.1 (-0.5) & 13.2 (-25\%) & 385 ($\times$1.3)\\
        \hline
        ViT-L$^{\text{MAE}}$ & 224 & 85.9 & 61.6 & 93 \\
        \textbf{VoMix-L}$^{\text{MAE}}_{r=(5\%)^{12}}$ & 224 & 85.3 (-0.6) & 40.2 (-35\%) & 137 ($\times$1.5) \\
        \hline
        ViT-H$^{\text{MAE}}$ & 224 & 86.9 & 167.4 & 36 \\
        \textbf{VoMix-H}$^{\text{MAE}}_{r=(5\%)^{12}}$ & 224 & 86.5 (-0.4) & 104.0 (-38\%) & 57 ($\times$1.6) \\
        \hline
        ViT-B@384 & 384 & 85.3 & 55.5 & 92 \\
        \textbf{VoMix-B}$_{r=(5\%)^{12}}^{@384}$ & 384 & 85.1 (-0.2) & 40.5 (-27\%) & 118 ($\times$1.3) \\
        \hline
        ViT-L@512 & 512 & 88.1 & 362 & 14.8 \\
        \textbf{VoMix-L}$_{r=(6\%)^{12}}^{@512}$ & 512 & 87.6 (-0.5) & 223 (-38\%) & 23.3 ($\times$1.6) \\
        \hline
        ViT-H@518 & 518 & 88.5 & 1017 & 5.2 \\
        \textbf{VoMix-H}$_{r=(6\%)^{12}}^{@518}$ & 518& 88.2 (-0.3) & 538 (-47\%) & 10.4 ($\times$2.0) \\
        \bottomrule
    \end{tabular}
    \caption{Evaluation results of ViT with VoMix on ImageNet-1K. ViT-X${^{\text{MAE}}}$ are the officially fine-tuned MAE models \cite{he2022masked} and ViT-B@384, ViT-L@512 and ViT-H@518 are released by SWAG \cite{singh2022swag}.}
    \label{tab:results_imagenet}
    \vspace{- 0.1 in}
\end{table*}

\subsection{Complexity Analysis}

We conduct a complexity analysis of VoMix to explore the additional time complexity.
Here, $N$ denotes the initial number of tokens in each layer, $D$ is the dimension of the feature representation, $H$ denotes the number of attention heads, and $r$ is the pruning ratio.

\textbf{Token Vote.} The complexity of head-wise mean of keys is $O(ND)$. Calculating the cosine similarity matrix $\mathbf{A}$ incurs $O(N^2D/H)$, and the voting complexity is $O(N^2)$. Given that $ D/H > 1 $, the dominant term is $O(N^2D/H)$ .

\textbf{Token Mix.} The complexity for soft-maxing weights is $O(N^2r(1-r))$, and for the query mix is $O(N^2Dr(1-r))$. Hence, the stepwise complexity is  $O(N^2Dr(1-r))$.

Aggregating the above components yields a total additional time complexity for VoMix of $O(N^2D(1/H + r(1-r)))$, which does not exceed  $O(N^2D)$. 
% Considering the progressively token pruning, the incremental complexity of VoMix is negligible compared to its benefits. 

\section{Experiments}

In this section, to verify the effectiveness of VoMix across different visual modalities, we conduct experiments on both image and video classification tasks. The experimental datasets are common benchmarks in these tasks: ImageNet-1K \cite{deng2009imagenet}, Kinetics-400 (K400) \cite{kay2017kinetics}, and Something-Something-V2 (SSV2) \cite{goyal2017something}. We apply VoMix to off-the-shelf models to re-evaluate their accuracy and speed, thereby verifying the plug-and-play capability of VoMix. All throughput results are obtained on a single 32GB Nvidia Tesla V100 GPU with a batch size of 32.

\textbf{Pruning Schedule.} VoMix is a token reduction method that relies on a hyperparameter $r^l$ to control the pruning ratio at the $l$-th layer. In our experiments, we set the value of $r$ for each layer to manage the tradeoff between accuracy and speed. We define two pruning schedules as follows:
\begin{itemize}
    \item constant schedule: $r=(a)^b$ indicates pruning a constant proportion of $a$ tokens in each of the first $b$ layers.
    \item decreasing schedule: $r=(a\downarrow)^b$ indicates pruning a decreasing proportion from $a$ to $0$ in the first $b$ layers.
\end{itemize}

\subsection{Image Experiments}
\label{sec:image}

We evaluate VoMix with several ViT models including MAE \cite{he2022masked}, SWAG \cite{singh2022swag} and DeiT \cite{touvron2021deit} on ImageNet-1K. We apply VoMix to the officially released fine-tuned models 
% \footnote{https://github.com/facebookresearch/mae} \footnote{https://github.com/facebookresearch/SWAG} \footnote{https://github.com/facebookresearch/deit}
to verify its effects on off-the-shelf models.

\begin{table}[t]
    \small
    \centering
    \begin{tabular}{lcc}
        \toprule
        Model & Acc & GFLOPs \\
        \hline
            \textcolor{black!50!white}{DeiT-S} \cite{touvron2021deit} & \textcolor{black!50!white}{79.8} & \textcolor{black!50!white}{4.6}  \\
            HVT-S-1 \cite{pan2021hvt} & 78.0 & 2.4 \\
            IA-RED$^{2}$ \cite{pan2021ia} & 78.6 & 2.9 \\
            A-ViT \cite{meng2022adavit} & 78.6 & 2.9 \\
            DynamicViT \cite{rao2021dynamicvit} & 79.3 & 2.9 \\
            SP-ViT \cite{kong2022spvit} & 79.3 & 2.6 \\    
            EViT \cite{liang2021evit} & 79.5 & 3.0 \\
            BAT \cite{long2023bat} & 79.6 & 3.0 \\
        \hline
        \rowcolor{blue!10!white}
        \textbf{VoMix-S}$_{r=(10\%\downarrow)^{12}}^{\text{DeiT}}$ & 78.6 & 2.9 \\
        \rowcolor{blue!10!white}
        \textbf{VoMix-S}$_{r=(12.5\%)^4}^{\text{AugReg}}$ & 79.5 & 2.9 \\
        \rowcolor{gray!10!white}
        \textbf{VoMix-S}$_{r=(10\%\downarrow)^{12}}^{\text{DeiT}}$ & \textbf{79.6} & 2.9 \\
        \bottomrule
    \end{tabular}
    \caption{Comparison on ImageNet-1K with other token reduction methods. Gray area means finetuned, while blue means without training.}
    \label{tab:compare_imagenet1k}
    
\end{table}

\begin{table}[t]
    \centering
    \begin{tabular}{lcc}
        \toprule
         Model & Acc & GFLOPs \\
         \hline
         \textcolor{black!50!white}{DeiT-S} \cite{touvron2021deit} & \textcolor{black!50!white}{79.8} & \textcolor{black!50!white}{4.6}  \\
         DeiT-S + ATS$^\dag$ \cite{fayyaz2022adaptive} & 76.9 & 2.5 \\
         DeiT-S + \textbf{VoMix}$_{r=(17\%)^{4}}$ & \textbf{77.3} & 2.5 \\
         \hline         
         DeiT-S + ATS$^\ddag$ \cite{fayyaz2022adaptive} & 72.7 & 2.0 \\
         DeiT-S + \textbf{VoMix}$_{r=(15\%)^{12}}$ & \textbf{75.4} & 2.0 \\
         \bottomrule
    \end{tabular}
    \caption{Comparison with a pluggable method ATS \cite{fayyaz2022adaptive}. We selected two tiers of GFLOPS, 2.5 and 2.0, respectively, to compare the performance of ATS and VoMix under plug-and-play conditions. \dag\ddag: from \citet{fayyaz2022adaptive} with the setting of \textit{Stage 3 Not Finetuned}.}
    \label{tab:compare_ats}
    
\end{table}

\begin{figure}[t]
    \centering
    \begin{subfigure}{0.9\linewidth}
         \includegraphics[scale=0.45]{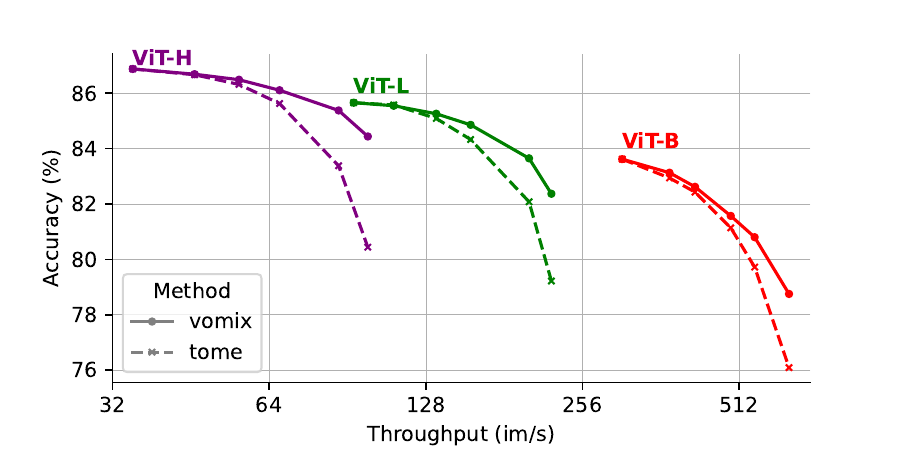}
        \caption{The speed-accuracy tradeoff of VoMix and ToMe.}
        \label{fig:mae_curve}
    \end{subfigure}
    \begin{subfigure}{0.9\linewidth}
        \includegraphics[scale=0.45]{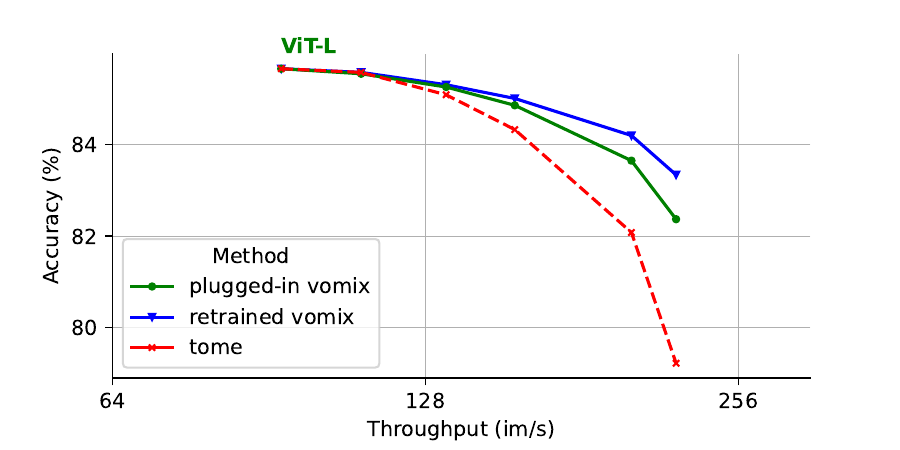}
        \caption{The speed-accuracy tradeoff of retrained VoMix-L$^\text{MAE}$ for 300 epochs.}
        \label{fig:train_curve}
    \end{subfigure}
    \caption{The speed-accuracy tradeoff on MAE models. We use the same pruning ratio settings for each method on the same tier of ViTs for fairness. The pruning values are $r=(3\%)^{12}, (5\%)^{12}, (7\%)^{12}, (10\%)^{12}, (12\%)^{12}$.}
    
\end{figure}

\textbf{Evaluation Results.} Table \ref{tab:results_imagenet} presents the acceleration effects of VoMix on various tiers and input resolutions of ViTs on ImageNet-1K. With an acceptable accuracy drop ranging from 0.2\% to 0.6\%, VoMix notably enhances the throughput for all tiers of ViTs. Larger ViTs exhibit a greater acceleration benefit. This is attributed to the fact that larger ViTs have deeper layers, which amplifies the cumulative effect of token reduction. In terms of input resolution, larger-sized inputs experience better acceleration with less precision loss, aligning with the intuition that high-resolution images contain higher redundancies. 

\textbf{Comparison with token reduction methods.} In Table \ref{tab:compare_imagenet1k}, we compare VoMix with several token pruning methods including HVIT \cite{pan2021hvt}, IA-RED$^{2}$ \cite{pan2021ia}, A-ViT \cite{meng2022adavit}, DynamicViT \cite{rao2021dynamicvit}, SP-ViT \cite{kong2022spvit}, EViT \cite{liang2021evit} and BAT \cite{long2023bat}. All these methods for comparision require retraining or further fine-tuning. By directly applying VoMix to DeiT-S \cite{touvron2021deit} without any training, we achieve the same accuracy and efficiency as A-ViT. We also apply VoMix on ViT-S-AugReg \cite{steiner2021augreg}, achieve accuracy comparable to other state-of-the-art methods with improved efficiency. It is noteworthy that VoMix does not require training, thereby actually saving training time. Futhermore, for a fair comparison, we  fine-tune VoMix from DeiT-S for 100 epochs, achieving results consistent with BAT \cite{long2023bat}. This indicates that VoMix not only achieves impressive results as a plug-and-play method but also has potential that can be further unlocked through training.

\textbf{Comparison with plug-and-play methods.} To evaluate the plug-and-play performance of VoMix, we make a comparison between VoMix and other pluggable token reduction methods. First, we compare VoMix with token merge (ToMe) \cite{bolya2022tome} on MAE models, and plot the tradeoff curves in Figure \ref{fig:mae_curve}. In the same configuration, VoMix presents a more favorable speed-accuracy tradeoff compared with ToMe. Specifically, at lower pruning ratios, the difference in accuracy is quite marginal; however, when the pruning ratio is further increased, ToMe suffers a significantly greater precision loss than VoMix. We hypothesize that this difference arises from the distinct pruning manners: ToMe merges token features in a hard manner, resulting in the combination of dissimilar tokens into one when many tokens are pruned. In contrast, VoMix selects queries through a voting mechanism and re-assigns feature information via a soft approach, thereby more effectively preserving the original features even with fewer tokens retained. Futhermore, Figure \ref{fig:train_curve} shows VoMix can be trained to get better performance. Additionally, we also compared VoMix with another pluggable method, ATS \cite{fayyaz2022adaptive}. Due to the requirement of ATS for ViT with a class token, our comparison is limited to DeiT. As is shown in Table \ref{tab:compare_ats}, with the same FLOPs cost, VoMix achieves higher accuracy when both two models are not fine-tuned.

\begin{figure}[t]
    \centering
    \includegraphics[scale=0.25]{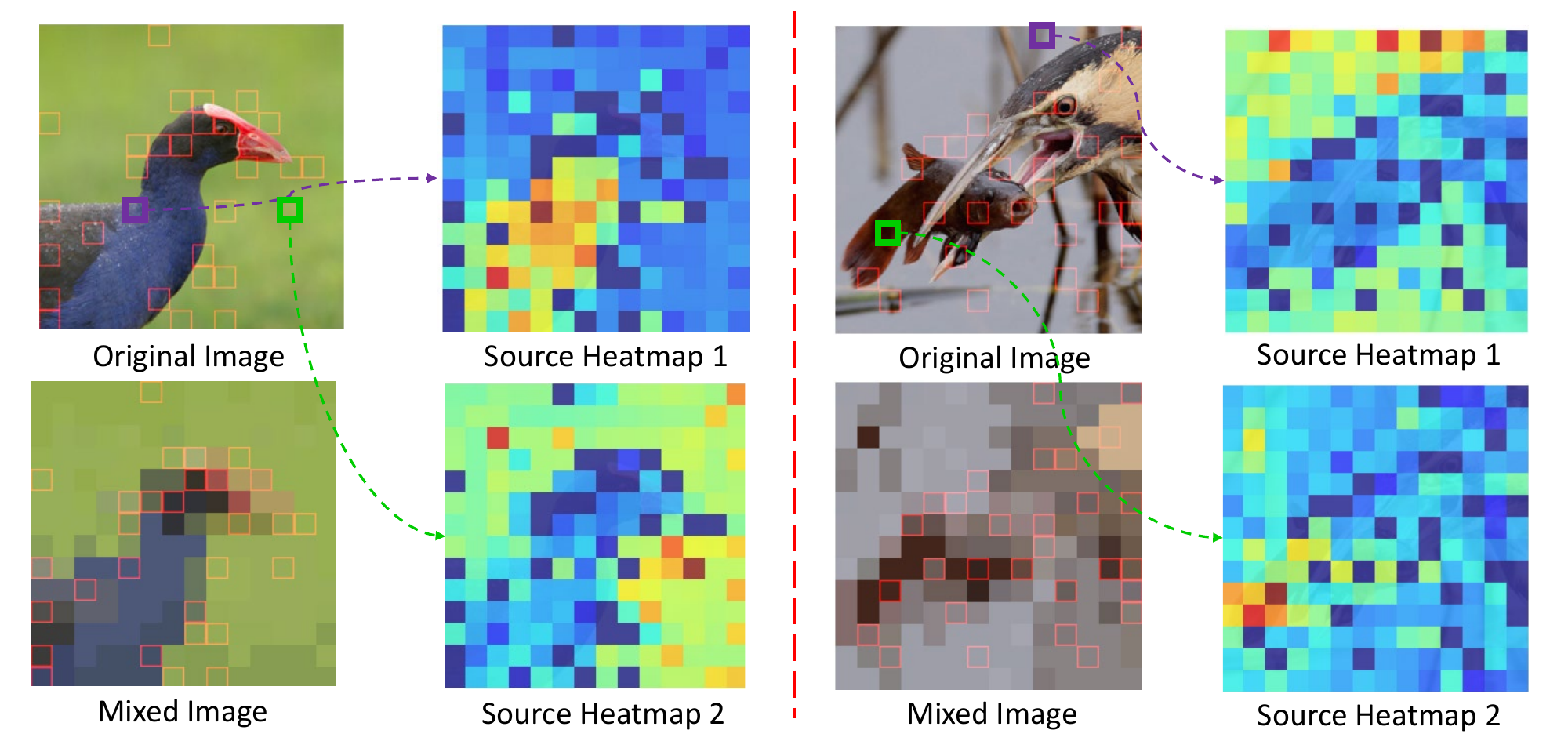}
    \caption{Visualization of feature source. The red fine boxes denote the final retained tokens by VoMix. The same color block in mixed image denotes they are primarily mixed into one token in the last layer. For each image, we select two representative tokens and visualize their feature source.}
    \label{fig:mixture_map}
    
\end{figure}

\textbf{Visualization.} To investigate how VoMix mixes token features, we visualize the tokens of the last layer and their source distribution in Figure \ref{fig:mixture_map} using ViT-L$_{r=(15\%)^{12}}$ on ImageNet-1K. We aim to address two key inquiries: (1) Which tokens does VoMix tend to retain? (2) From which tokens do the retained tokens draw information?

\begin{figure}[t]
    \centering
    \includegraphics[scale=0.25]{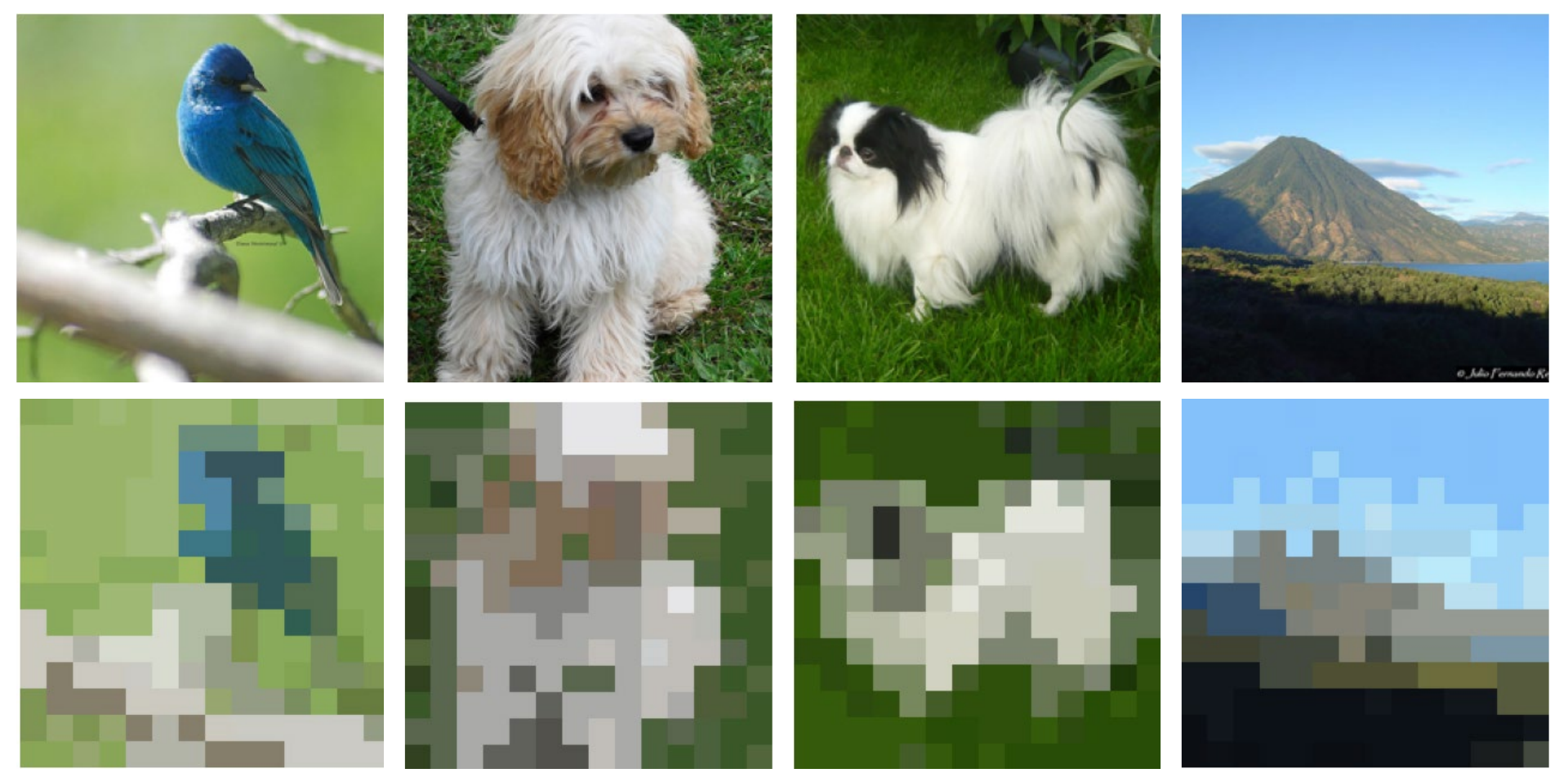}
    \caption{Image Visualization. The two rows display the original images and the mixed images. The color blocks indicate that VoMix mixes the region into one token.}
    \label{fig:image_vis}
    
\end{figure}

For the first inquiry, we find that unlike previous pruning methods that only retain foreground tokens, VoMix preserves at least one representative token for each semantic region. More tokens are retained in semantic-rich regions, like the bird's head, with fewer tokens for the background region. Moreover, the retained tokens are strategically placed at the boundaries of semantic regions, highlighting VoMix's capability to prioritize dissimilar tokens, thereby emphasizing edge tokens as excellent representatives. This mechanism encourages the model to focus on contour features, steering away from redundancy within the interior of regions.

Addressing the second inquiry, we elucidate the feature sources of the retained tokens by selecting two tokens from each image and visualizing their source heatmaps. These heatmaps, where hotter areas indicate higher feature weights being mixed into the selected token, reveal the diverse source distribution of different retained tokens. In the left image, the bird's nape (purple box) primarily draws features from its body, while the grass token (green box) mainly draws from the background. In the right image, the fish's tail (green box) mainly derives its features from its tail fin and the water area token (purple box) from the background. This pattern of feature aggregation demonstrates VoMix's functionality akin to token clustering, where it aggregates similar token features around a retained token, reducing redundancy by merging similar tokens into representative regions. 

These findings are further supported by the visualizations in Figure \ref{fig:image_vis}, which make it apparent that VoMix tends to cluster similar tokens into the same region, thereby substantiating our analysis of how VoMix mixes token features to achieve efficient and effective representation.

\subsection{Video Experiments}
\label{sec:video}

We conducted experiments on two video classification datasets: Kinetics-400 (K400) \cite{kay2017kinetics} and Something-Something-V2 (SSV2) \cite{goyal2017something}, using VideoMAE \cite{tong2022videomae} as the base model. We apply VoMix to the officially released fine-tuned models 
% \footnote{https://github.com/MCG-NJU/VideoMAE}
and conduct evaluation.

\textbf{Video Clip} Considering the need to segment videos into clips for video experiments, we adopt the clip settings of VideoMAE \cite{tong2022videomae} for fairness. During the evaluation, we sample 5 clips $\times$ 3 crops with 16 frames for K400 and 2$\times$3 views for SSV2. For throughput evaluation, we report the throughput of 16-frame 224$\times$224 clips per second. 

\begin{table}[t]
    \small
    \centering
    \setlength{\tabcolsep}{2.5pt}
    \begin{tabular}{lllll}
        \toprule
         \multirow{2}{*}{Model} & \multicolumn{2}{c}{Acc} & \multirow{2}{*}{GFLOPs} & \multirow{2}{*}{clip/s} \\
        & K400 & SSV2 & & \\
         \midrule
         ViT-S & 79.1 & 66.8 & 57 & 66.4 \\
        \textbf{VoMix-S}$_{r=(5\%)^{12}}$ & 78.9 & 66.5 & 40 (-30\%) & 73.6 ($\times1.1$) \\
         \midrule
         ViT-B &  81.5 & 70.5 & 180 & 24.7 \\
         \textbf{VoMix-B}$_{r=(5\%)^{12}}$ & 81.3 & 70.6 & 128 (-29\%) & 31.9 ($\times1.3$) \\
         \textbf{VoMix-B}$_{r=(30\%\downarrow)^{12}}$ & 80.2 & 68.0 & 60 (-67\%) & 67.6 ($\times2.7$) \\
         \midrule
         ViT-L & 85.2 & - & 597 & 8.4 \\
         \textbf{VoMix-L}$_{r=(9\%)^{12}}$ & 85.0  & - & 249 (-58\%) & 19.8 ($\times2.4$) \\
         \textbf{VoMix-L}$_{r=(12\%)^{12}}$ & 84.6 & - & 195 (-67\%) & 25.3 ($\times3.0$) \\
         \midrule
         ViT-H & 86.4 & - & 1192 & 4.9 \\
         \textbf{VoMix-H}$_{r=(7\%)^{12}}$ & 86.1 & - & 567 (-52\%) & 9.5 ($\times1.9$) \\
         \bottomrule
    \end{tabular}
    \caption{Evaluation results of ViT with VoMix on K400. All the models are pretrained by VideoMAE. VoMix can scale larger ViTs to the same throughput as the low-tier but obtain higher accuracy.}
    \label{tab:results_k400}
    
\end{table}

\begin{table}[t]
    \small
    \centering
    \setlength{\tabcolsep}{3pt}
    \begin{tabular}{lcc}
        \toprule
        Model & Acc & GFLOPs$\times$Views \\
        \midrule
        VideoSwin-B \cite{liu2022video} & 82.7 & 338$\times$10$\times$5  \\
        \textcolor{black!50!white}{ViT-L$^{\text{MAE}}$} \cite{tong2022videomae} & \textcolor{black!50!white}{85.2} & \textcolor{black!50!white}{597$\times$5$\times$3}  \\
        ToMe-ViT-L$^{\text{MAE}}$ \cite{bolya2022tome} & 84.5 & 281$\times$10$\times$1  \\
        STA-ViT-L$^{\text{MAE}}$ \cite{ding2023sta} & 85.0 & 308$\times$5$\times$3  \\
        \midrule
        \textbf{VoMix-ViT-L}$^{\text{MAE}}_{r=(9\%)^{12}}$ & 85.0 & \textbf{249}$\times$5$\times$3 \\
        \toprule
        % TimeSformer-L \cite{bertasius2021timesformer} & 80.7 & 8353$\times$1$\times$3  \\
        Motionformer-L \cite{patrick2021motionformer} & 80.2 & 1185$\times$1$\times$3  \\
        VideoSwin-L \cite{liu2022video} & 84.9 & 2107$\times$10$\times$5  \\
        MViTv2-L \cite{li2022mvitv2} & 86.1 & 2828$\times$1$\times$3  \\
        \textcolor{black!50!white}{ViT-H$^{\text{MAE}}$} \cite{tong2022videomae} & \textcolor{black!50!white}{86.4} & \textcolor{black!50!white}{1192$\times$5$\times$3}  \\
        ToMe-ViT-H$^{\text{MAE}}$ \cite{bolya2022tome} & 86.1 & 609$\times$5$\times$3  \\
        STA-ViT-H$^{\text{MAE}}$ \cite{ding2023sta} & 86.1 & 611$\times$5$\times$3  \\
        \midrule
        \textbf{VoMix-ViT-H}$^{\text{MAE}}_{r=(7\%)^{12}}$ & 86.1 & \textbf{567}$\times$5$\times$3 \\
        \bottomrule
    \end{tabular}
    \caption{Comparisons with state-of-the-art method on K400.}
    \label{tab:comparison_k400}
    
\end{table}

\begin{table}[t]
    \small
    \centering
    \begin{tabular}{lll}
        \toprule
        model & acc & GFLOPs$\times$views  \\
        \midrule
        TimeSformer-L & 62.4 & 5549$\times$1$\times$3 \\
        Motionformer-L & 68.1 & 1185$\times$1$\times$3 \\
        STTS-Swin-B & 68.7 & 237$\times$1$\times$3 \\
        VideoSwin-B & 69.6 & 321$\times$1$\times$3 \\
        MViTv2-B & 70.5 & 225$\times$1$\times$3 \\
        \textcolor{black!50!white}{ViT-B$^{\text{MAE}}$} & \textcolor{black!50!white}{70.5} & \textcolor{black!50!white}{180$\times$2$\times$3}  \\
        STA-ViT-B$^{\text{MAE}}$ & 70.3 & 116$\times$2$\times$3  \\
        \midrule
        \textbf{VoMix-ViT-B}$^{\text{MAE}}_{r=(5\%)^{12}}$ & \textbf{70.6} & 128$\times$2$\times$3 \\
        \bottomrule
    \end{tabular}
    \caption{Comparisons with state-of-the-art method on SSV2.}
    \label{tab:comparison_ssv2}
\end{table}

\textbf{Evaluation Results} Table \ref{tab:results_k400} shows the results of ViT with VoMix on K400 and SSV2. Starting from ViT-B, we report two results in the table: one with a slight loss in accuracy, and the other with throughput comparable to the lower tier ViT. With only a 0.2\% $\sim$ 0.3\% decrease in accuracy, VoMix reduces the computational cost by approximately 30\% for low-tier ViTs (ViT-S, ViT-B) and 60\% for high-tier ViTs (ViT-L, ViT-H). The actual throughput increase aligns closely with the reduction in computational cost, demonstrating the additional computational cost introduced by VoMix is negligible compared to its benefits. By further increasing the pruning ratio, VoMix achieves a dual advantage in both accuracy and speed for the high-tier ViT over the low-tier one. Figure \ref{fig:intro_performance} shows the improvement of speed-accuracy tradeoff introduced by VoMix.

\begin{figure}[t]
    \centering
    \includegraphics[scale=0.55]{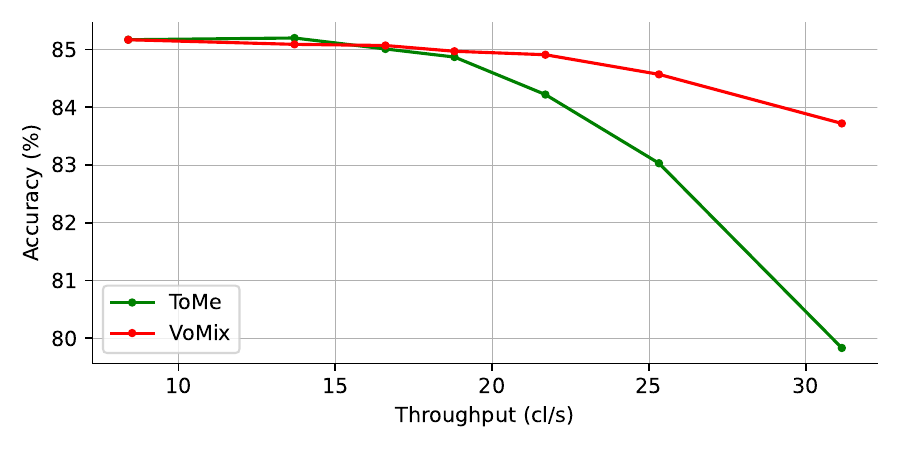}
    \vspace{-0.2in}
    \caption{The speed-accuracy tradeoff of VoMix and ToMe on K400 using ViT-L$^{\text{MAE}}$ with the same pruning ratios of $r=(5\%)^{12}, (7\%)^{12}, (8\%)^{12}, (10\%)^{12}, (12\%)^{12}, (15\%)^{12}$.}
    \label{fig:k400_curve}
    
\end{figure}

\textbf{Comparison with State of the Art} We compare VoMix with other state-of-the-art work on K400 and report the results in Table \ref{tab:comparison_k400}. The results are manually split into two tracks according to the FLOPs range. We include video-specific models like TimeSformer \cite{bertasius2021timesformer}, Motionformer \cite{patrick2021motionformer}, VideoSwin \cite{liu2022video}, MViTv2-L \cite{li2022mvitv2} and two pluggable token pruning methods based on VideoMAE \cite{tong2022videomae} models: ToMe \cite{bolya2022tome} and STA \cite{ding2023sta} as the baselines. In both two tracks, VoMix outperforms other models in terms of accuracy and computational cost. ViT with VoMix significantly surpasses video-specific models in both accuracy and speed. Compared with two pluggable pruning methods, VoMix achieves the same accuracy with less computational cost. Furthermore, we completely compare the speed-accuracy tradeoff between VoMix and ToMe on K400 using ViT-L$^{\text{MAE}}$ in Figure \ref{fig:k400_curve}. Similar to the results on ImageNet-1K, ToMe is slightly ahead at lower pruning ratios. However, as the pruning ratio increases, ToMe suffers a highly significant loss in accuracy while VoMix maintains a better accuracy.

\begin{figure}[t]
    \centering
    \includegraphics[scale=0.4]{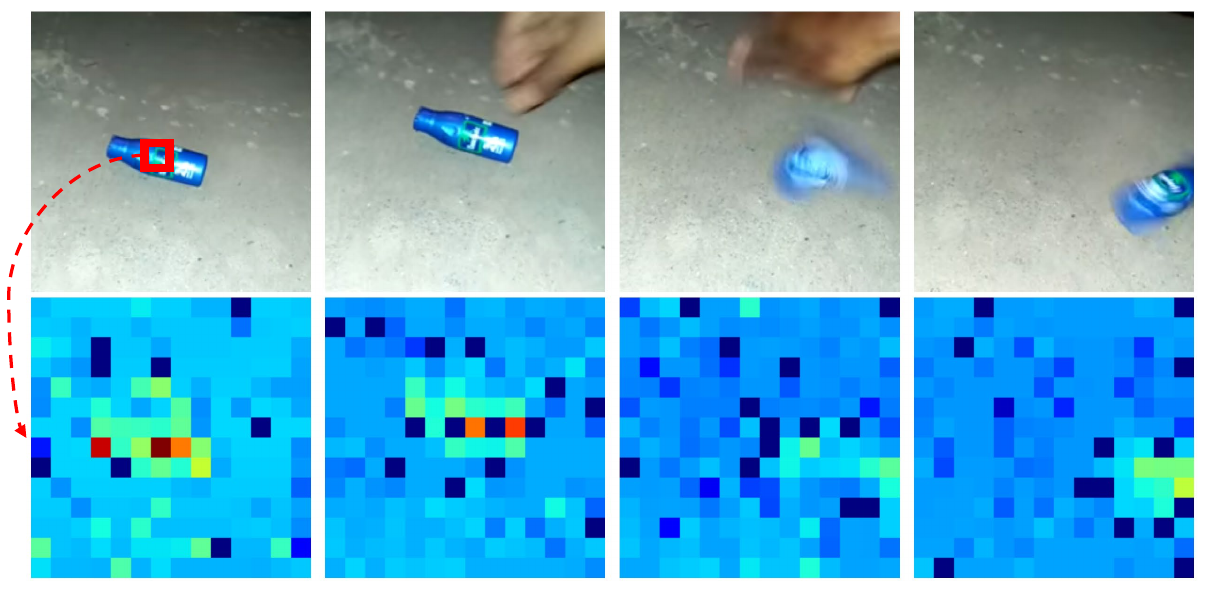}
    \caption{Video Visualization. The two rows display the video clip and source heatmap of the red-boxed token.}
    \label{fig:video_vis}
\end{figure}

\textbf{Visualization} \label{para:vis} Similar to image visualization, we visualize the source heatmap over multiple frames of video using VoMix-L$^{\text{MAE}}_{r=(40\%\downarrow)^{12}}$ in Figure \ref{fig:video_vis}. We select a final retained token (red box) of the blue bottle and track the mixture source. As is shown in the heatmap, it mainly draws features from the blue bottle across the frames, which indicates that VoMix can also perform feature aggregation on video.

\subsection{Ablation Study}
\label{ablation}

To investigate the optimal strategy, we conduct ablation studies on ImageNet-1K using ViT-L@512 from SWAG \cite{singh2022swag}. The results are displayed in Table \ref{tab:ablation}.

\textbf{Selection Strategy} Three strategies include (1) \textbf{voting strategy} employed by VoMix; (2) \textbf{global maximum similarity}, which selects tokens with the highest average similarity to all the other tokens; (3) \textbf{random selection}, which randomly selects tokens. Compared to global similarity, voting strategy demonstrates a clear advantage. This is attributed to the locality of voting, meaning that the selected tokens are not required to be globally most similar, but only to exhibit the highest similarity among several tokens.

\textbf{Voting Mechanism} To explore how many tokens should a token vote to, we examine three settings: (1) vote for \textbf{top 1}; (2) vote for \textbf{top 2}; (3) vote for \textbf{top $r$}. Top 1 outperforms others, supporting the aforementioned conclusion that the superiority of voting strategy lies in its locality.

\begin{table}[t]
    \small
    \centering
    \begin{subtable}{.3\linewidth}
    \centering
        \begin{tabular}{ll}
            Strategy & Acc  \\
            \midrule
            \rowcolor{gray!50!white}
            vote & \textbf{87.54} \\
            max sim & 87.17 \\
            random & 86.90 \\ 
        \end{tabular}
        % \caption{Selection Strategy.}
    \end{subtable}
    \begin{subtable}{.3\linewidth}
    \centering
        \begin{tabular}{ll}
            Vote & Acc \\
            \midrule
            \rowcolor{gray!50!white}
            top 1 & \textbf{87.54} \\
            top 2 & 87.47 \\
            top r & 87.42 \\ 
        \end{tabular}
        % \caption{Voting Mechanism.}
    \end{subtable}
    \begin{subtable}{.3\linewidth}
    \centering
        \begin{tabular}{ll}
            Feature & Acc\\
            \midrule
            q & 87.46 \\
            \rowcolor{gray!50!white}
            k & \textbf{87.54} \\
            v & 87.45 \\ 
        \end{tabular}
        % \caption{Similarity Metrics.}
    \end{subtable}
    
    \begin{subtable}{.3\linewidth}
    \centering
        \begin{tabular}{ll}
            Similarity & Acc  \\
            \midrule
            \rowcolor{gray!50!white}
            cosine & \textbf{87.54} \\
            L2 dist & 87.28 \\
            dot & 87.26 \\ 
        \end{tabular}
        % \caption{Similarity Evaluation.}
    \end{subtable}
    \begin{subtable}{.3\linewidth}
    \centering
        \begin{tabular}{ll}
            Q-Mix & Acc  \\
            \midrule
            \rowcolor{gray!50!white}
            global & \textbf{87.54} \\
            max & 87.33 \\
            no mix & 87.39 \\ 
        \end{tabular}
        % \caption{Query Mix.}
    \end{subtable}
    \begin{subtable}{.3\linewidth}
    \centering
        \begin{tabular}{ll}
            Attn-Mix & Acc \\
            \midrule
            \rowcolor{gray!50!white}
             mix & \textbf{87.54} \\
             no prop & 87.48  \\
             no mix & 87.17 \\ 
        \end{tabular}
        % \caption{Attention Mix.}
    \end{subtable}
    
    \caption{Ablation studies on ImageNet-1K of ViT-L@512 with $r=(7\%)^{12}$. \colorbox{gray!50!white}{gray} indicates the default settings.}
    \label{tab:ablation}
    
\end{table}

\begin{figure}[t]
    \centering
    \includegraphics[scale=0.55]{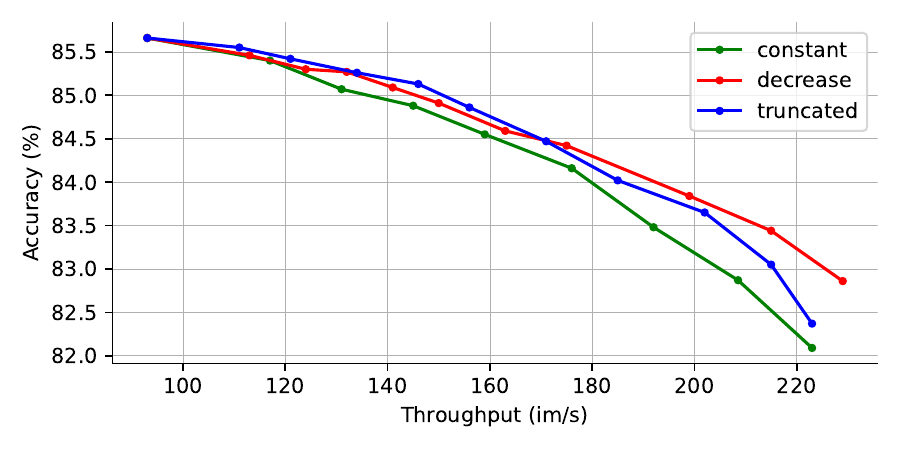}
    \vspace{-0.2in}
    \caption{Pruning schedules of ViT-L$^{\text{MAE}}$ on ImageNet-1K, denoted as $r=(a)^{24}$, $r=(a\downarrow)^{24}$, $r=(a)^{12}$.}
    \label{fig:pruning_schedule}
    
\end{figure}

\textbf{Similarity Measurement} We utilize three features to measure similarity: \textbf{q}, \textbf{k}, \textbf{v}. Using \textbf{k} as the metric performs best. Besides, we experiment with three methods of similarity measurement: cosine similarity, L2 distance, and vector dot product. Cosine similarity outperforms others in similarity measurement.

\textbf{Query Mix} We explore the effects of three different query mixing strategies: (1) \textbf{global mix}, where the selected queries are mixed according to the similarity to all retained queries; (2) \textbf{max mix}, where the selected queries are mixed only with the most similar retained query; (3) \textbf{no mix}, where the selected queries are discarded without any mixing. The global query mix outperforms the others, indicating the superiority of soft-manner mixing.

\textbf{Attention Mix} We explore the effects of attention mix with the three settings: (1) \textbf{attention mix} employed by VoMix, which performs proportion attention with retained \textbf{q} and original \textbf{k},\textbf{v}; (2) \textbf{no proportion attention}; (3) \textbf{no mix}, where ViT performs attention with retained \textbf{q}, \textbf{k}, \textbf{v}. The results show no mixing suffers a significant precision loss, indicating that after query mixing, attention should be performed with the full set of keys and values.

\begin{table}[t]
    \small
    \centering
    \begin{tabular}{llccc}
    \toprule
        Train setting & Infer setting & Acc & im/s & hours \\
        \hline
        default & default & 85.87 & 93 & 26 \\
        \hline
        default & VoMix$_{r=(5\%)^{12}}$ & 85.26 & 137 & 26 \\
        VoMix$_{r=(5\%)^{12}}$ & default & 85.73 & 93 & 18 \\
        VoMix$_{r=(5\%)^{12}}$ & VoMix$_{r=(5\%)^{12}}$ & 85.31 & 137 & 18 \\
    \bottomrule
    \end{tabular}
    \caption{Training ViT-L$^\text{MAE}$ on ImageNet-1K applying VoMix on 8 V100 GPUs for 300 epochs.}
    \label{tab:train_results}
    \vspace{-0.2in}
\end{table}

\subsection{Discussion}

\textbf{Pruning Schedule} We design three pruning schedules: (1) constant schedule: a constant proportion of tokens are pruned across all layers; (2) decreasing schedule: the pruning ratio gradually decreases to zero across layers; (3) truncated schedule: pruning is performed only at the early half layers. The results are illustrated in Figure \ref{fig:pruning_schedule}. The constant schedule is almost the worst strategy at any throughput. At lower pruning ratios, the truncated schedule performs better, while at higher ratios, the decreasing schedule surpasses it.

\textbf{Should I train VoMix?} We have demonstrated the potential of training VoMix in Table \ref{tab:compare_imagenet1k} and Figure \ref{fig:train_curve}. Here, we further discuss the time and performance benefits brought by training VoMix. We train ViT-L$^{\text{MAE}}$ applied VoMix from scratch on ImageNet-1K using the fine-tuning scripts of MAE \cite{he2022masked}. Results are shown in Table \ref{tab:train_results}. Training with VoMix results in a slight increase in accuracy compared with plug-and-play mode. Notably, training with VoMix and inferring on vanilla ViT-L only suffers 0.1\% accuracy drop but saves nearly 30\% training time. It indicates that training VoMix further enhances the accuracy-speed tradeoff, and also effectively speeds up training.

\section{Conclusion}

In this work, we introduce Vote\&Mix (VoMix), a plug-and-play and parameter-free token reduction method, which can be readily applied to off-the-shelf ViT models \textit{without any training}. VoMix tackles computational redundancy of ViTs by voting and mixing tokens with high homogeneity. Experiments demonstrate that VoMix significantly improves the speed-accuracy tradeoff of ViTs on both images and videos and surpasses the existing token reduction methods.

\bibliography{aaai25}

\end{document}